\documentclass{article}

\usepackage[numbers]{natbib}

\usepackage[final]{neurips_2020}
\usepackage{amsmath}
\usepackage[utf8]{inputenc} 
\usepackage[T1]{fontenc}    
\usepackage{hyperref}       
\usepackage{url}            
\usepackage{booktabs}       
\usepackage{amsfonts}       
\usepackage{nicefrac}       
\usepackage{microtype}      
\usepackage{makecell}
\usepackage{graphicx}
\usepackage{multirow}
\usepackage{bm}
\usepackage{colortbl}
\usepackage{amsthm}         
\usepackage{caption}
\usepackage{subcaption}
\usepackage[export]{adjustbox}

\title{Monitoring the Impact of Wildfires on Tree Species with Deep Learning}

\author{%
  Wang Zhou \\
  IBM T.J. Watson Research Center \\
  IBM Research\\
  Yorktown Heights, New York 10598\\
  \texttt{wang.zhou@ibm.com} \\
   \And
   Levente Klein \\
   IBM T.J. Watson Research Center \\
   IBM Research\\
   Yorktown Heights, New York 10598\\
   \texttt{kleinl@us.ibm.com} \\
}

\usepackage[textsize=scriptsize]{todonotes}
\usepackage[normalem]{ulem}

\begin{document}

\maketitle

\begin{abstract}
One of the impacts of climate change is the difficulty of tree regrowth after wildfires over areas that traditionally were covered by certain tree species. 
Here a deep learning model is customized to classify land covers from four-band aerial imagery before and after wildfires to study the prolonged consequences of wildfires on tree species. 
The tree species labels are generated from manually delineated maps for five land cover classes: Conifer, Hardwood, Shrub, ReforestedTree and Barren land. With an accuracy of 92\% on the test split, the model is applied to three wildfires on data from 2009 to 2018. The model accurately delineates areas damaged by wildfires, changes in tree species and regrowth in burned areas. 
The result shows clear evidence of wildfires impacting the local ecosystem and the outlined approach can help monitor reforested areas, observe changes in forest composition and track wildfire impact on tree species.
\end{abstract}

\section{Introduction}

In the last decades the frequency, the intensity and the damage caused by wildfires is increasing in the Pacific Northwest \cite{wildfire_usa}.  
While wildfires can cause tremendous economic and social losses, one of their prolonged consequences is the resulting change of the vegetation species \cite{Fairman, Barrett, Magadzire}. With climate change, the temperature and soil moisture create unfavorable conditions for vegetation regrowth in areas that were devastated by wildfires. The higher temperatures and lesser soil moisture impact the germination of seeds in locations where in the past they could strive. For example, multiple studies \cite{tree_regeneration,pine_climate} have demonstrated that  rose pines and blue oaks may not regenerate in their traditional locations. The climate change may affect the variety and density of trees in areas affected by wildfires, and thus it is important to systematically monitor the long-term tree species distributions in fire-prone regions.

Classification of tree species is often carried out by forest agencies based on visual inspection of aerial imagery \cite{SierraNevada_2011}.  These kinds of vegetation identification campaigns are sparse in time with irregular updates, which may not be sufficient to register the changes in vegetation caused by wildfires happening almost yearly. 

Here we propose a deep learning based classification method to monitor the impact of wildfires on tree species. 
Large-scale image classification on multispectral images to detect tree species before and after wildfires can quantify the impact of wildfires on vegetation species. 
We apply the method to three historical wildfire regions in California and the results show clear changes in vegetation.
With high resolution images readily available from aerial or satellite observations, tracking tree species across large areas over extended period of time is possible. Monitoring vegetation in near real time can be used by forest services and environmental agencies to better plan for forest management after natural disaster events and quantify ecological disasters.

\section{Related Work}

The impact of climate change on vegetation shift in North West USA has been studied since 1990s \cite{pine_climate, vegclima}. 
Droughts, wildfires and pests inflicted noticeable impact on trees health in Sierra Nevada Mountains, California \cite{SN_trees}. 
To quantify the impact of wildfires on tree regeneration, field surveys \cite{Fairman, Debouk, Cai} are often conducted but limited to only a few sample sites.
Hansen \emph{et al.} \cite{hansen} plant small plots of trees in areas with significant environmental differences and observe that some of the tree species cannot germinate or survive under changed climate with increased ambient temperature and soil moisture, which suggests that climate change will rearrange certain tree species habitat. However, the controlled planting studies are labor intensive and rely on continuous supervision on the plots, which restricts the scale of such studies.  

Quick classification of trees from aerial and satellite imagery can enable large-scale survey of tree species and track climate impact on their survival in their traditional habitats.  Convolutional neural networks have been used to identify forests/trees in satellite snapshots \cite{chimera,deepsum,autogeo}, dense time series of imagery \cite{SR_sentinel}, and hyperspectral imagery \cite{hyperspectral}. Tracking tree species before and after wildfires was not pursued in any of the above studies. Additionally Lidar point clouds are classified to recognize tree species \cite{lidar_tree} or combination of hyperspectral data and Lidar \cite{hyper_lidar}, but access to high resolution Lidar scans is not readily available for most of the locations on the globe.

\section{Datasets and Method}

\paragraph{Datasets.} The  National Agricultural Imagery Program (NAIP) collects high resolution aerial imagery in four spectral bands (Red, Green, Blue and Near Infrared) every other year for the last decades \cite{naip}. The spatial resolution for the most recent acquisitions is 0.6 m while older images are acquired at 1 m resolution. The imagery is collected in full leaf season, offering a consistent way to compare the vegetation status year to year. In this study, NAIP images from 2009, 2012, 2014, 2016 and 2018 are analyzed. All the data are resampled to the same spatial resolution (0.6 m) as part of the data processing.

\paragraph{Labels.} Labeled vegetation data is extracted from manual labels created in 2011 for the Sierra Nevada mountains \cite{SierraNevada_2011}. Manually delineated polygons contain vegetation classes \emph{Conifer}, \emph{Hardwood},  \emph{Shrub} and urban areas. 
For the tree covered regions the density of each tree class is specified. The classes were filtered based on the coverage density to identify locations with a specific tree species. 
Additionally, two more classes were grouped, \emph{ReforestedTree}, where newly planted trees were sampled to identify the characteristics of reforested regions, and \emph{Barren}, where all the identified tree classes of \cite{SierraNevada_2011} were small.
Each polygon was then separated in non-overlapping areas of the size of $32\times32$ pixels and signed with the label from the annotated polygons.
The associating NAIP data within the identified areas were extracted from PAIRS Geoscope platform \cite{pairs}. 
The data were further filtered by eliminating samples which had low Normalized Difference Vegetation Index (NDVI) \cite{ndvi} values and were most likely not encapsulating vegetation information for Conifer and Hardwood \cite{kleindim}. Table~\ref{tab:tree_data} lists the number of samples for the curated dataset. In total, 93,849 samples are collected for training, and two sets of 5,000 samples for validation and testing, respectively.

\begin{table}[phb]
    \centering
    \caption{Dataset statistics of the tree species.}
    \vspace{2mm}
    \label{tab:tree_data}
    \begin{tabular}{l c r}
         \hline
         Tree type & Label & \# points \\
         \hline
		Conifer           & 0 &  18,708 \\
		Hardwood         & 1 & 19,873 \\
		Shrub            & 2 & 24,430 \\
		ReforestedTree          & 3 &  21,701 \\
		Barren             & 4 & 19,137 \\
         \hline
         Total              & & 103,849 \\
         \hline
    \end{tabular}
\end{table}

\paragraph{Wildfires.} Wildfire boundaries were obtained from California \cite{wildfire_datasource} and analyzed to investigate the variability of tree species before and after wildfires. Specifically, Swedes Fire in 2013 and Wall Fire in 2017 both of which happened in Butte County, CA, and Fletcher Fire from 2007 in Modoc County, CA were reported. These regions are not covered by the training samples.

\paragraph{Networks.} A modified version of ResNet34 \cite{resnet} is used for the classification of tree species. The network was specially changed to accommodate the four-channel input data compared to regular three-channel RGB images. Since the training data is noisy and limited, smoothed labels $y^{LS}_k$ \cite{inception3, smooth2, smooth} were used to compute a CrossEntropy loss instead of hard one-hot labels $y_k$,
\begin{equation}
    y^{LS}_k = y_k (1-\alpha) + \alpha/K,
\end{equation}
where $K=5$ is the total number of classes and $\alpha$ is the label smoothing factor, in order to mitigate over-fitting and extreme gradients from wrong labels.

\section{Experiments}

\paragraph{Configurations.} 
Our experiment setup is as follows. For training, an SGD optimizer with a momentum of 0.9 and a weight decaying of 0.0005 is used. 
The learning rate is set to be 0.1 and divided by 10 every 100 epochs, and the model is trained for 300 epochs in total with a mini-batch size of 512. Label smoothing factor $\alpha$ is set to be 0.1. Random horizontal/vertical flipping, rotation and random cropping is applied at training, while for testing no data augmentation is used.
For large areas of interest, the data are diced into $32\times32$ pixel tiles, and fed batch by batch to the network for classification at testing. The classification results are then assembled to recreate a classification map. A $3\times3$ majority filter is applied on the classification map to reduce noise.
Our implementation is based on Pytorch.

\begin{figure}[!hpb]
    \centering
    \includegraphics[width=0.9\textwidth]{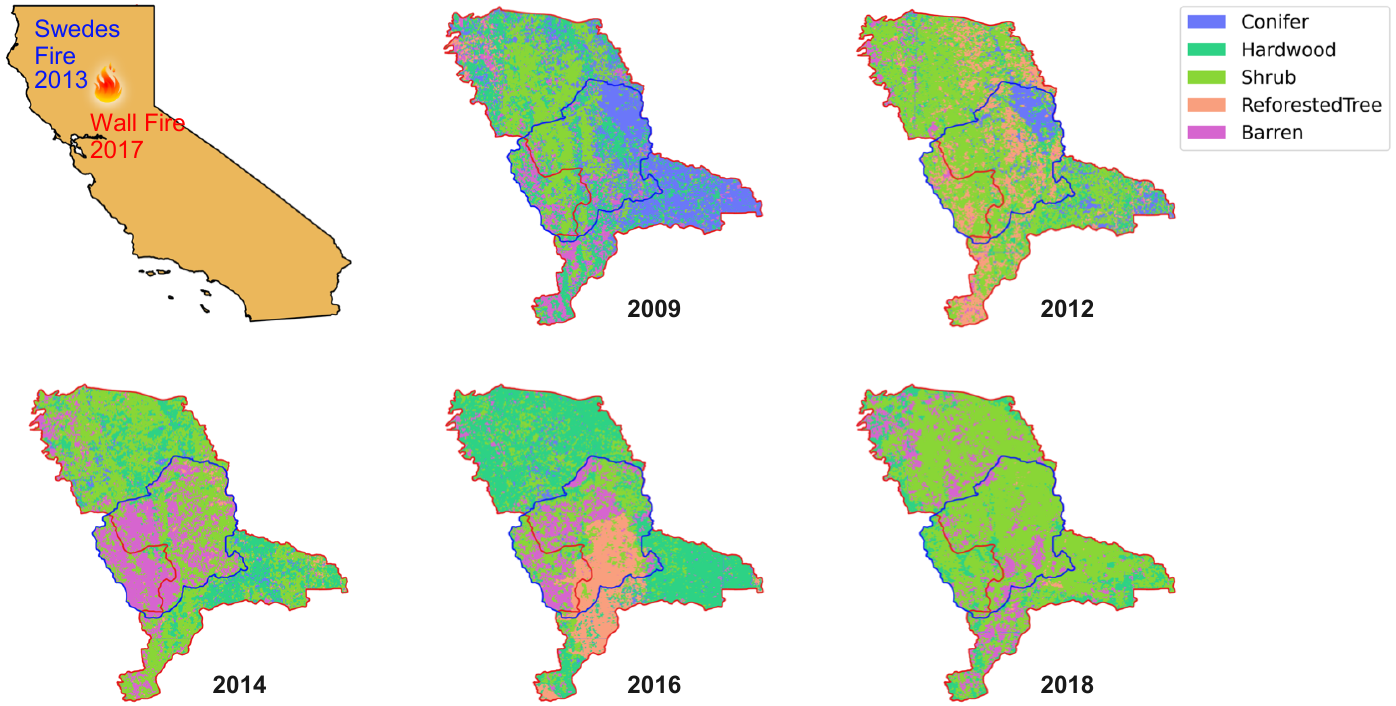}
    \caption{Tree species classification of burned regions of 2013 Swedes Fire (blue outline) and 2017 Wall Fire (red outline) in Butte County, CA.}
    \label{fig:WF4}
\end{figure}

\begin{figure}[!tb]
    \newcommand{\figwidth}{0.48}
	\centering
	\begin{subfigure}[t]{\figwidth\columnwidth}
		\centering
		\includegraphics[width=1\columnwidth]{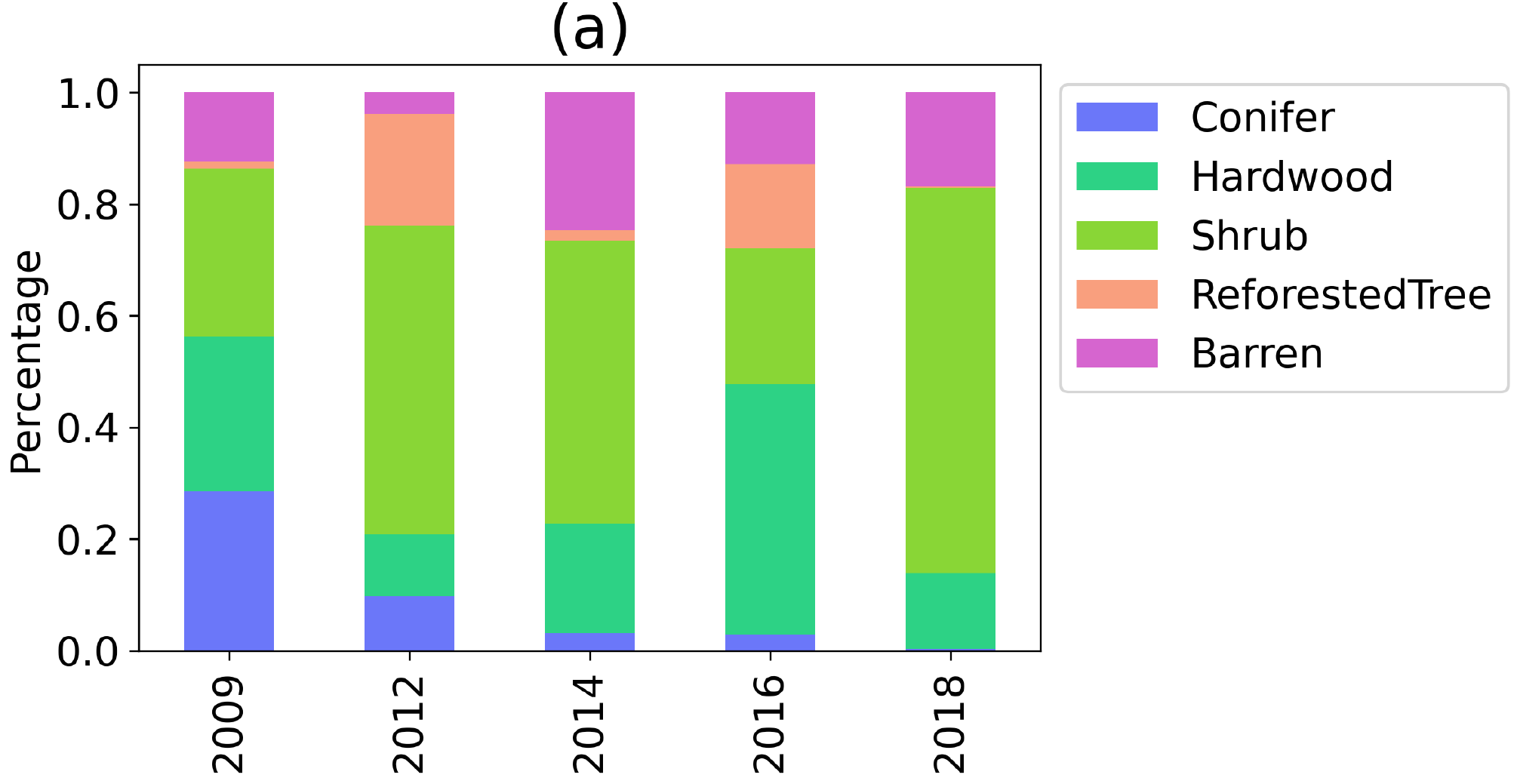}
	\end{subfigure}
	\begin{subfigure}[t]{\figwidth\columnwidth}
		\centering
		\includegraphics[width=1\columnwidth]{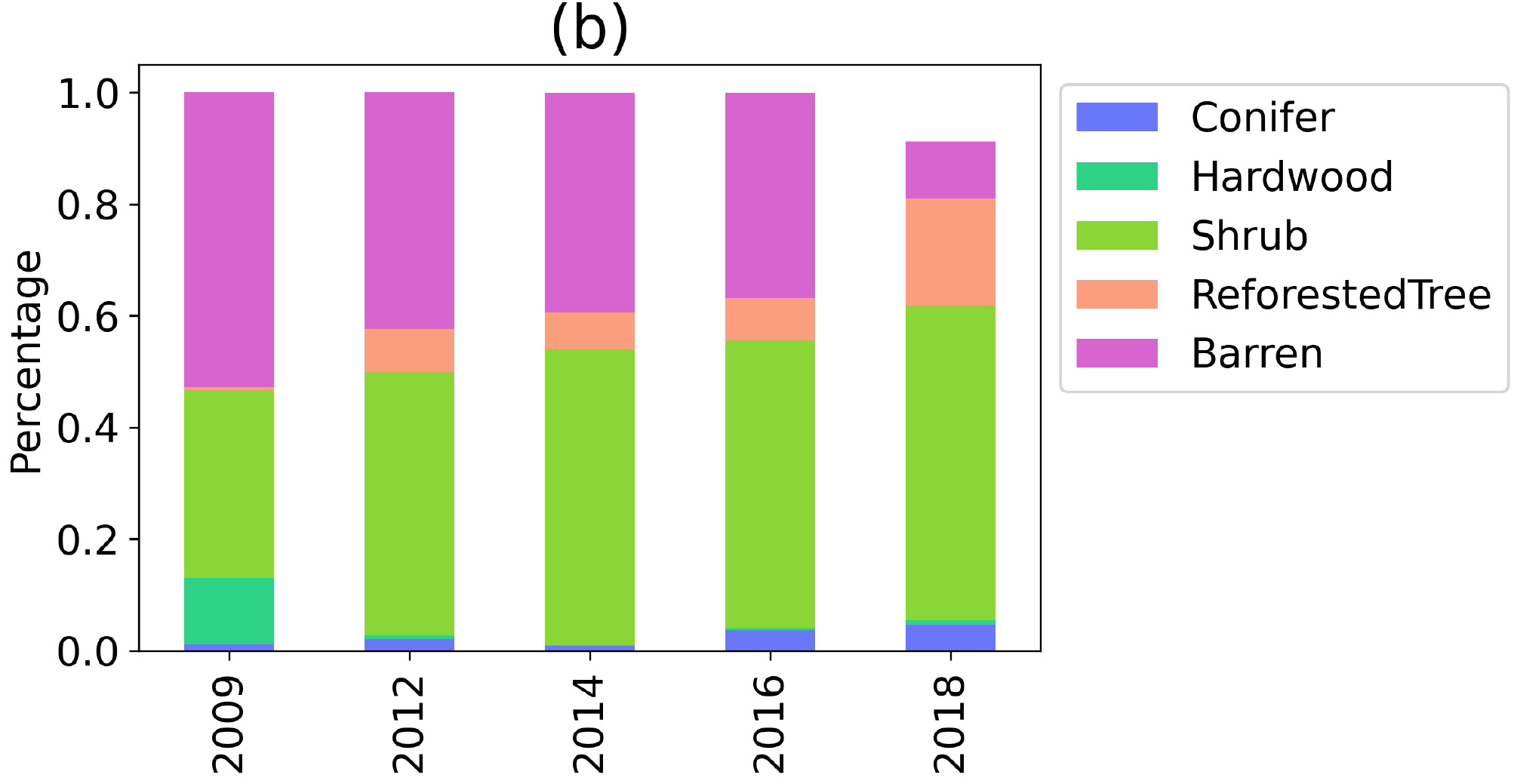}
	\end{subfigure}
    \caption{Land cover distribution of (a) 2013 Swedes Fire and 2017 Wall Fire together and (b) 2007 Fletcher Fire (incomplete bar for 2018 is due to missing data in Oregon State in 2018).}
    \label{fig:stats}
\end{figure}

\begin{figure}[!tb]
    \centering
    \includegraphics[width=0.9\textwidth]{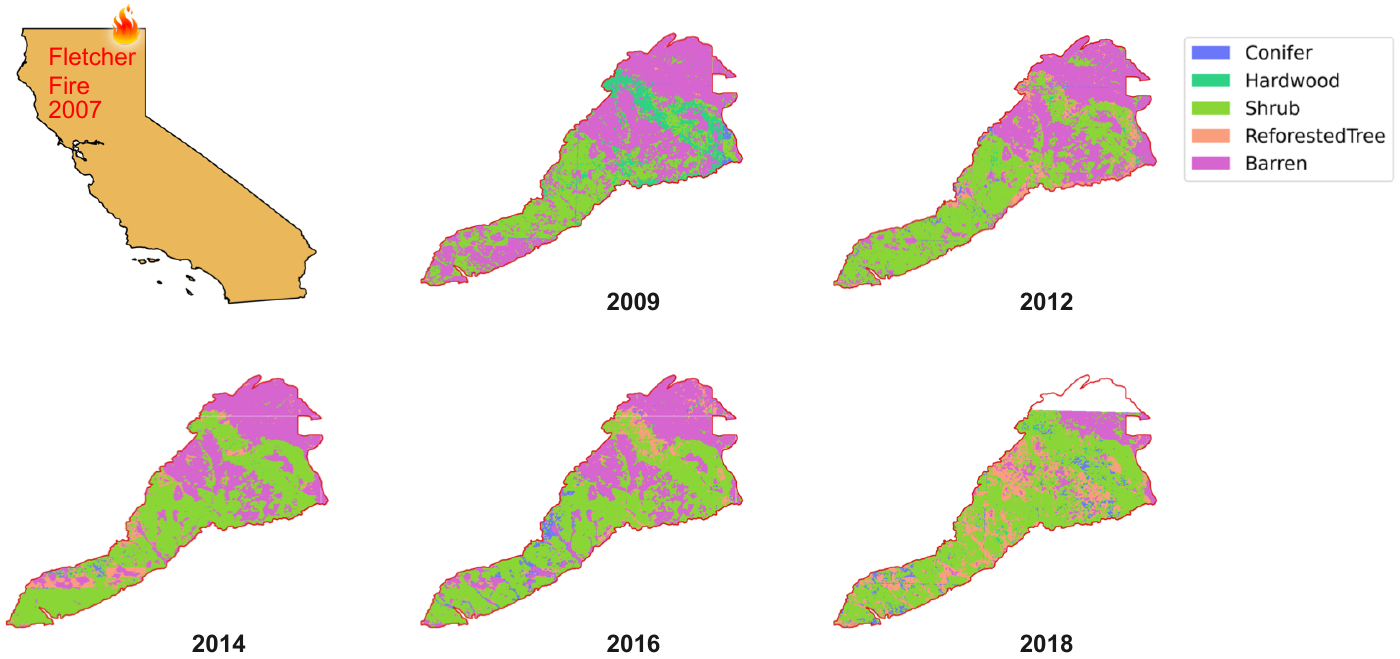}
    \caption{Tree species classification of burned regions of 2007 Fletcher Fire in Modoc County, CA. The missing part of the 2018 classification is due to missing data in Oregon State in 2018.}
    \label{fig:WF3}
\end{figure}

\paragraph{Results.} The model is first evaluated on the test split of the curated dataset. The overall classification accuracy is $\mathbf{92.2\%}$ on the test data. The model is then applied to three wildfire regions to generate the classification maps in order to study the changes in tree species in those affected regions. Figures~\ref{fig:WF4}~and~\ref{fig:WF3} depict the tree species maps across different years, and in Figure~\ref{fig:stats} the bar plots illustrate the area distributions of each class normalized by the total area.

In Figure~\ref{fig:WF4}, two of the wildfires were studied, 2013 Swedes Fire marked with blue outline and 2017 Wall Fire marked with red outline\footnote{In November 2018, the disastrous Camp Fire started in the same region. NAIP data of 2018 were acquired before the wildfire, and therefore it reflects the vegetation status before the Camp Fire.}. Comparing the map for the blue-bordered region between 2012 and 2014, a large patch of the vegetated areas was cleared by the Swedes Fire and turned into bare land as indicated by the Barren class. The bar plot (Figure~\ref{fig:stats}a) shows a sudden expansion of Barren land areas after 2012, with 18\% of the region converting to no vegetation covered land. In 2016, half of the bare land area from 2014 was covered by new vegetation and Hardwood coverage is doubled by extending into the previously Shrub area. A new disruption occurred in 2017, caused by the Wall Fire. Most of the Hardwood as well as ReforestedTree disappeared, and without the regrowth of trees, Shrub expanded by almost three times. It is evident from Figure~\ref{fig:WF4} and Figure~\ref{fig:stats}a that frequent wildfires hurt the regrowth of trees, and forest areas may be permanently removed and covered by grass and shrubs, which are more prone to potential fires.

Interestingly, there is a consistent trend of decline of Conifer in this area, which is captured by the decline of area percentage of Conifer in Figure~\ref{fig:stats}a over the years. This may reflect the slow decline of conifer trees at the foothill of Sierra Nevada, as is also observed in \cite{SN_trees}. Our approach can serve as a tool to study long-term changes in tree species at large scales. While this study covers only the last decades of forest composition change due to data availability, it reveals some consistent trends that can be observed by the current climate impact on California's forests.

Since the 2007 Fletcher Fire took place before any of the NAIP data were collected, there are no abrupt changes in the species in this area (Figure~\ref{fig:WF3}). Due to the fire, half of the land remained bare in 2009. However, there has been uninterrupted regrowth of trees across the area, with an increasing distribution of ReforestedTree observed in Figure~\ref{fig:stats}b. The area of Barren is  decreasing as being converted to trees and shrub regrowth. The trend suggests that if there were no further wildfires, this area can recover and sustain the vegetation on long term.

\section{Conclusion}
A deep learning model is trained to classify tree species from aerial imagery and track the changes in tree species before and after three wildfires in California over a 9 year period. 
The model tracks multiple major tree species and  validates that some tree species vanish from the areas affected by the wildfire. The model accurately recognizes long-term changes in areas that were reforested to preserve forest composition. This approach can be used to monitor the forest composition across large geographical areas in response to climate change, enabling foresters and environmental groups to make more informed decisions to preserve ecosystem balance. 

\clearpage
\section*{Broader Impact}
Climate change is increasing the frequency and the intensity of wildfires across the globe, causing tremendous loss of lives and economic damages. Besides larger areas being burned, the climate change is impeding tree regrowth in areas that were covered prior to wildfire, which in turn accelerates the climate change.
Deep learning techniques are used to classify tree species in remote sensing images and track forest composition changes.  
Quantitative evaluation of tree density and tree species detection can help forest services, ecological organization and environmental groups to carry out comprehensive studies to preserve current forests and ensure the carbon reduction through reforestation.

{\small
\bibliographystyle{unsrt}
\bibliography{ref}}


\end{document}